\documentclass[conference]{IEEEtran}

\makeatletter
\newcommand{\linebreakand}{%
  \end{@IEEEauthorhalign}
  \hfill\mbox{}\par
  \mbox{}\hfill\begin{@IEEEauthorhalign}
}
\makeatother

\IEEEoverridecommandlockouts
\usepackage[T1]{fontenc} 
\usepackage[utf8]{inputenc}
\usepackage{cite}
\usepackage{amsmath,amssymb,amsfonts}
\usepackage{algorithmic}
\usepackage{graphicx}
\usepackage{textcomp}
\usepackage{xcolor}
\usepackage{booktabs}
\usepackage{hyperref}
\usepackage{multirow}
\DeclareUnicodeCharacter{2032}{'}
\def\BibTeX{{\rm B\kern-.05em{\sc i\kern-.025em b}\kern-.08em
    T\kern-.1667em\lower.7ex\hbox{E}\kern-.125emX}}

\begin{document}
\pdfoutput=1
\title{Towards a Sample Efficient Reinforcement Learning Pipeline for Vision Based Robotics}

\author{\IEEEauthorblockN{Pierre BELAMRI}
\IEEEauthorblockA{\textit{Research Intern CAOR} \\
\textit{MINES ParisTech, PSL Research University}\\
Paris, France \\
pierre.belamri@mines-paristech.fr}

\and

\IEEEauthorblockN{Maxence MAHÉ}
\IEEEauthorblockA{\textit{Research Intern CAOR} \\
\textit{MINES ParisTech, PSL Research University}\\
Paris, France \\
maxence.mahe@mines-paristech.fr} \\

\linebreakand

\IEEEauthorblockN{Jesùs BUJALANCE MARTIN}
\IEEEauthorblockA{\textit{CAOR} \\
\textit{ Mines ParisTech, PSL Research Univeristy}\\
Paris, France \\
jesus.bujalance\_martin@mines-paristech.fr}

}

\maketitle
\begin{abstract}
Deep Reinforcement learning holds the guarantee of empowering self-ruling robots to master enormous collections of conduct abilities with negligible human mediation. The improvements brought by this technique enables robots to perform difficult tasks such as grabbing or reaching targets. Nevertheless, the training process is still time consuming and tedious especially when learning policies only with RGB camera information. This way of learning is capital to transfer the task from simulation to the real world since the only external source of information for the robot in real life is video. In this paper, we study how to limit the time taken for training a robotic arm with 6 Degrees Of Freedom (DOF) to reach a ball from scratch by assembling a pipeline as efficient as possible. The pipeline is divided into two parts: the first one is to capture the relevant information from the RGB video with a Computer Vision algorithm. The second one studies how to train faster a Deep Reinforcement Learning algorithm in order to make the robotic arm reach the target in front of him. Follow this link to find videos and plots in higher resolution: \url{https://drive.google.com/drive/folders/1_lRlDSoPzd_GTcVrxNip10o_lm-_DPdn?usp=sharing}
\end{abstract}


\section{Introduction}
Today, technological advances have allowed robots to manipulate objects and perform tasks just as well or even better than humans\cite{zhang_towards_2015}. This is evidenced by the increasing automation of assembly plants and even the use of robotic arms on Mars rovers. The applications are almost infinite, from surgical medicine to the dismantling of nuclear power plants and the handling of radioactive waste. However, programming these robots is tedious and time consuming. In addition, they cannot adapt to a new environment which limits their usage and their interaction with humans. Therefore, one of the research challenges is to make the collaborative robots as adaptive as possible. It is necessary that the behavior of these robots is not defined in a fixed way and written by a programmer. 

\begin{figure}[htbp]
\centerline{\includegraphics[width=0.5\textwidth]{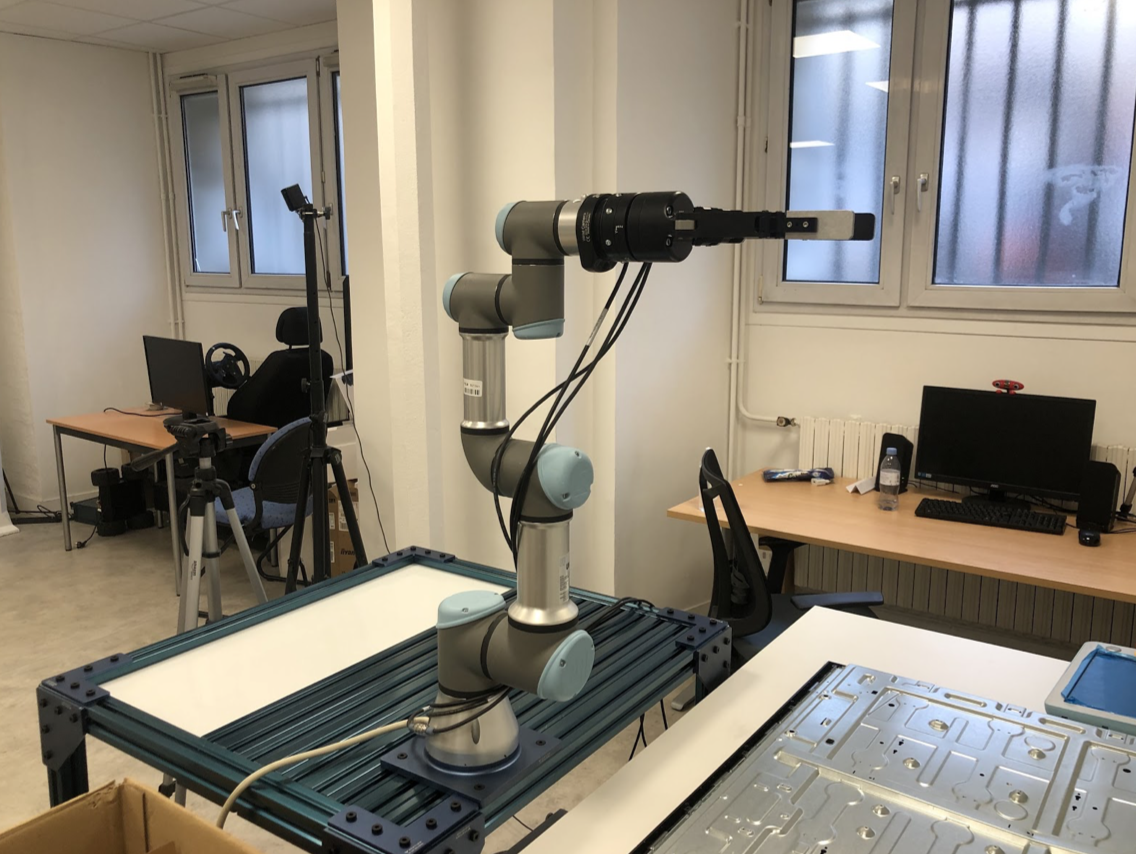}}
\caption{The UR3 Robot in CAOR, Mines Paris.}
\label{fig}
\end{figure}

Recent breakthroughs especially in Deep Reinforcement Learning (DRL) have led to uncommon capacities in self-governing tasks such as reaching a target or even grabbing only square objects. Reinforcement Learning (RL) algorithms learn how to act by trial and error\cite{Thrun92efficientexploration}. By succeeding or failing, the agent gets short-terms rewards that add up to a long-term reward. The ultimate goal is to define the best sequence maximizing the long-term reward. DRL incorporates deep learning in the learning process allowing agents to make decisions from unstructured input data without manual engineering of the state space\cite{Fran_ois_Lavet_2018}. DRL has mastered and surpassed humans in well known classical games such as Chess and Go and video games such as Dota5\cite{slavkovik_deep_2019}. 

However, reinforcement learning methods need hundreds of hours of training and important computational power making them extremely sample inefficient. It's why, today, robotic agents learn their policy in a simulation environment such as RLBench in order to test millions of steps more easily\cite{Katyal_2017_CVPR_Workshops}, \cite{9001253}. This accelerates the training. Nevertheless, one of the drawbacks of simulation is a policy with a high variance, hardly transferable to tasks in the real world. Indeed, the robot learns inside a simulation environment where it can know its coordinates and the coordinates of the target in real time. Therefore it is able to locate itself in space and more easily learn how to reach. It is not possible to know this much information in real life training only with cameras and the joints' positions. 

In this paper, we propose a training pipeline using off policy RL algorithm to efficiently learn how to reach a target. Our pipeline is in three stages: 1) an imitation learning process. 2) a computer vision algorithm locates on the camera stream the target and the robot's head. 3) a DRL algorithm is trained based on these observations, to learn how to reach. This pipeline is easy to adapt to any reaching and grasping situation since it relies only on imitation learning and camera observation and is also easy to modify. 
In order to assemble the most efficient pipeline we had to study different configurations with different algorithms, this benchmark was an important part of our work and is detailed in this paper. In addition, for security purposes, we studied in  RLBench\cite{9001253}, a simulated environment in order to try different configurations and not to damage the real robot. The ultimate goal is to transfer what was learnt in the simulated environment, in real life with as few changes as possible. It is why our simulated environment was very close to reality with cameras placed around the robot in order to recreate the real world configuration. 

\section{Related Work}

\subsection{Deep Reinforcement Learning (DRL)}
Deep reinforcement learning overcame the difficulties encountered when dealing with high dimensional state-space where previous RL methods had issues to choose the most efficient task. Thanks to its neural network, DRL achieved state of the art results on difficult tasks such as self driving cars\cite{pan2017virtual} to poker game\cite{brown2017safe}. It is especially useful to learn tasks from pixel inputs such as in video games\cite{shao2019survey} or simulation\cite{9001253}. However the combination of off-policy learning and high-dimensional, nonlinearfunction approximation with neural networks presents a major challenge for stability and convergence. In this paper we studied one of the more recent off policy algorithms in DRL which is the Soft Actor Critic method\cite{haarnoja_soft_2018} that achieved state of the art results. The choice of an off-policy algorithm is because off-policy algorithm are more sample efficient for difficult tasks such as robotic manipulation.
\bigbreak
\textbf{Soft Actor Critic(SAC)}
The Soft Actor Critic algorithm is an off-policy method relying on two major pillars. The first one is the actor critic architecture with a separate policy and value function, the second is the maximization of the entropy in order to favor stability and exploration. 
Actor-Critic algorithms have an actor network and a critic network which updates the actor network with a predefined frequency\cite{morgan2021model}. In other words, it alternates between evaluation and improvement. Off policy actor-critic obtains faster training times than on-policy. However, it is very unstable and difficult to train\cite{degris2013offpolicy}. To answer this problem SAC introduces the soft actor critic by adding the maximization of the entropy to the algorithm in addition to maximizing the expected return. The result is better stability and faster training with state of the art results, regardless of the policy parameterization. 
\begin{figure}[htbp]
\centerline{\includegraphics[width=0.53\textwidth]{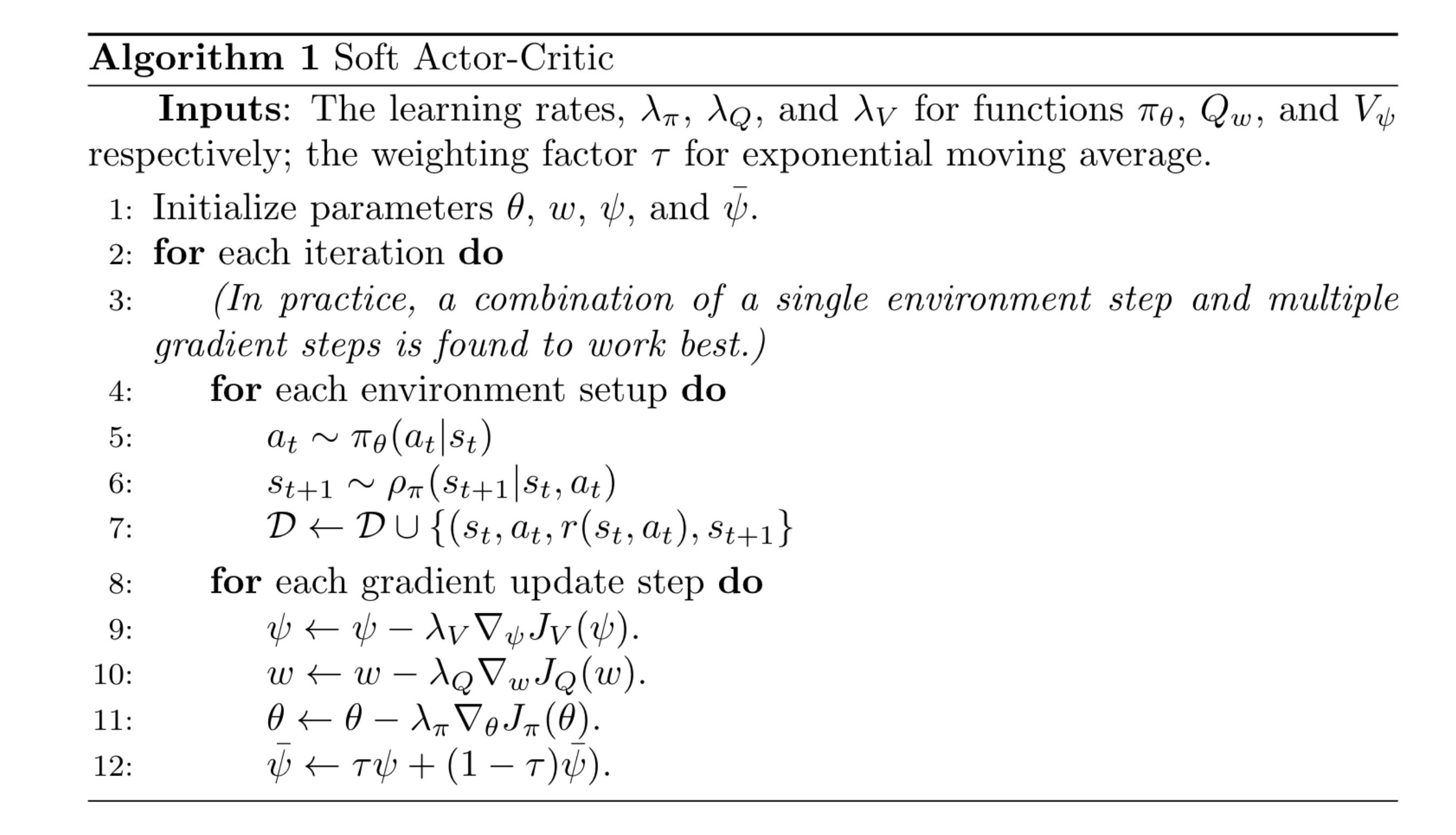}}
\end{figure}
\bigbreak
\textbf{Vision Based Reinforcement Learning}
Vision based RL for robotic tasks is a recent research subject. It was first studied in a relatively simple configuration with 3 Degrees Of Freedom (DOF) robot and in two dimensions\cite{zhang_towards_2015}. More recently researches have been using vision based RL to train robots with higher degrees of freedom and in three dimensions but a majority of experiments consists in using many robots in parallel to train which is costly and hard to reproduce\cite{kalashnikov_scalable_nodate}. Others have been focusing on training only in real life which can be difficult and may harm the robot if not done properly\cite{zhan_framework_2020}. Our method concentrates on the replicability of the experiment and the possibility to adapt it easily to other tasks. This is possible thanks to the architecture by blocks and with an upstream simulation training. 
\subsection{Imitation Learning}
Imitation Learning aims to mimic the behavior of an expert for a certain task. One way to leverage demonstration data for an off-policy DRL agent is to initialize the agent's memory with demos showing him how to succeed the task\cite{4399517}. Implementing imitation learning in a reinforcement learning process enables the agent to learn faster and therefore decreases the training time\cite{ren2020generalization}. The most natural form of imitation consists in creating a memory buffer of perfect behaviors and feed to the agent in order for him to take examples from it. In this paper, we use demos from the simulation environment. For each demo we record the visual representation and the proprioception of the robot (joints' positions). 
\subsection{Single Shot Detector}
Single Shot Detector (SSD) is a computer vision algorithm used for its capacity to process frames in real-time. Unlike Mask-R-CNN\cite{he2018mask} and Faster-R-CNN\cite{ren2016faster} running respectively at 5 fps and 7 fps, SSD can run at more than 50 fps without losing in precision\cite{liu_ssd_2016}. The backbone of the algorithm can change according to needs, in this paper, we use a MobileNetV1\cite{howard2017mobilenets} backbone in order to be implemented easily without needing high computational power. The SSD network feeds forward the image only once and uses the output feature maps of the backbone network as the inputs for its layers. In addition, SSD uses Atrous convolution\cite{chen2017rethinking} instead of conventional convolution which enables the network to have fewer number of parameters and therefore be faster.  
\subsection{Contrastive Learning}
Contrastive Learning (CL) is a machine learning technique used to extract high dimensional features from an unlabeled data set by learning which data points are similar or different\cite{islam2018deep}.
This self-supervised learning method works by maximizing a contrastive loss function if two pairs of images are similar and by minimizing it if they are different. It relies on data augmentation (crop, color distortion...) in order to create a positive pair. The images (positive and negative pairs) are fed to a neural network and a projection head. Then the similarity function tells if the images are similar\cite{chen2020big}. The loss function used in this paper is the InfoNCE loss\cite{oord2019representation}.

\section{Overview}

To provide an overview of our work, we have to begin introducing the reaching task, followed by the environment in which we are performing this task, and then, an overview of the pipeline will be presented, before studying in detail the different parts of it.

\subsection{Environment}

\begin{figure}[htbp]
\centerline{\includegraphics[width=0.4\textwidth]{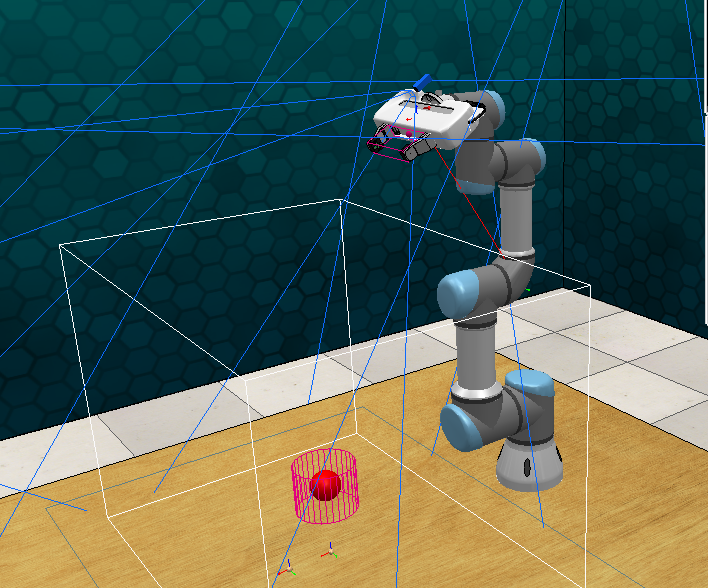}}
\caption{The UR3 robot within the CoppeliaSim simulator.}
\label{fig}
\end{figure}

CoppeliaSim \cite{coppeliaSim} is a simulator commonly used in the robotics community.
Within CoppeliaSim, there are a number of options for physics simulators, like Bullet. 
\begin{figure}[htbp]
\centerline{\includegraphics[width=0.5\textwidth]{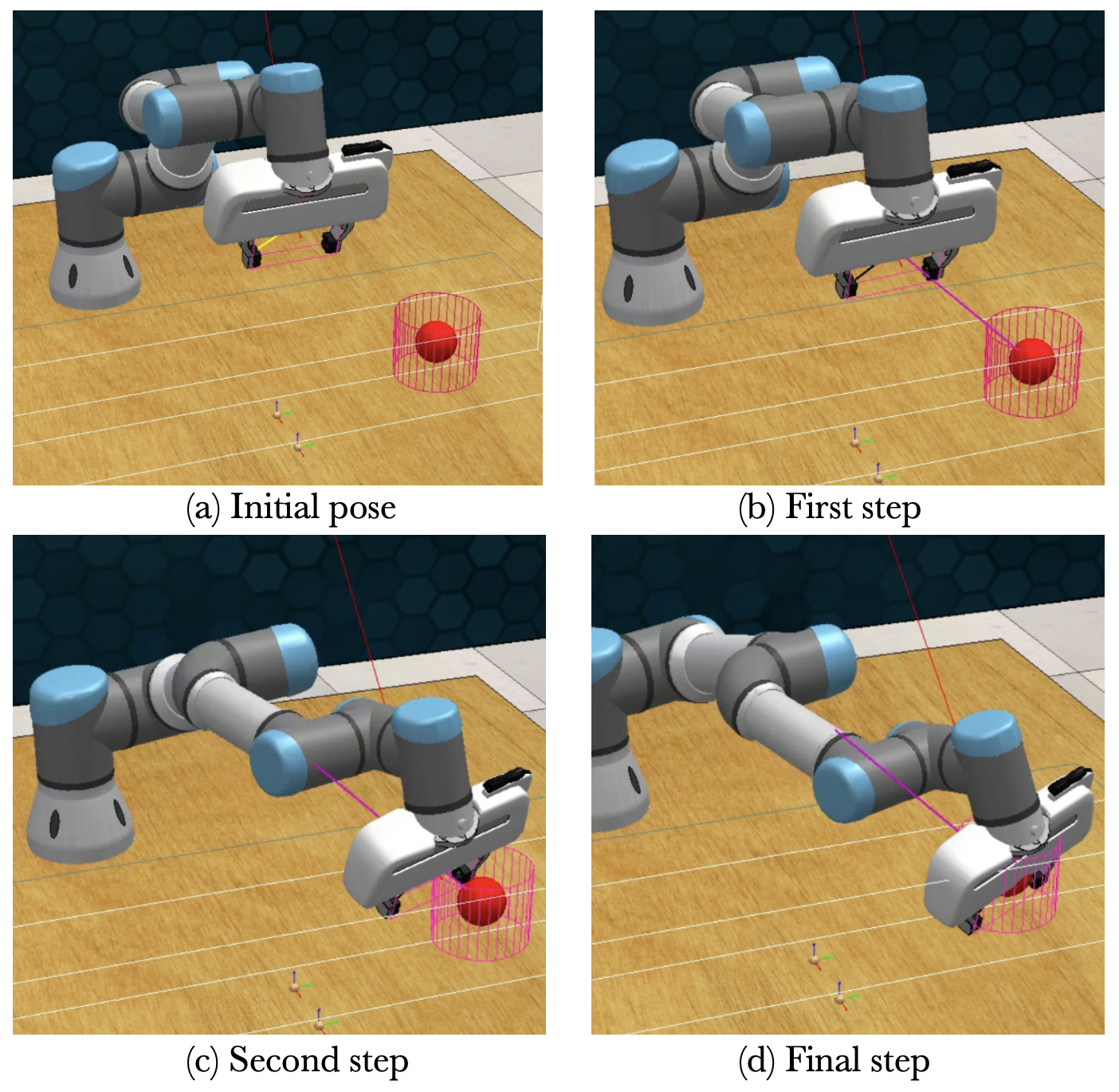}}
\caption{Three steps of the reaching task for UR3. The red ball represents the target pose. (a) The initial pose of the arm. (b-c) Robot positions after the first and second steps bringing the robot’s tip nearer to the target pose. (d) The ultimate success step: once the distance of the positions between the gripper and target are within the desired precision, the step is memorized as a successful move. 
The number of steps required for a pose reach might vary in several target poses.}
\label{fig}
\end{figure}
For this we decided to use and control it  with Python thanks to the PyRep \cite{james2019pyrep} toolkit, to which we added the learning environment RLBench \cite{9001253}, which offered an initial structure for reinforcement learning.
It is in this environment that we carried out the training on the reach task.
\subsection{Reach Task}
The reaching task is a basic element of robotic manipulation. It consists in the robot's gripper to reach the target while satisfying Cartesian space constraints. We aim to use reinforcement learning to train a UR3 on one goal attainment task, during which a policy function is trained to learn to map current states to the robot's next action to reach the target goal. An example of a trained goal attainment is shown in Fig.3.

Many recent studies have targeted position-only reaching  \cite{DBLP:journals/corr/abs-1709-07643} or situations wherever the tip effector is forced to reach the target from one direction, usually the upper vertical \cite{lenz2014deep} one. Such simplification will greatly simplify tasks however is tough to generalize to interact with objects that need to be reached in a certain way.

\subsection{Pipeline}

Before studying in detail the different parts of our training pipeline, her is an overview of it. 

\begin{figure}[htbp]
\centerline{\includegraphics[width=0.48\textwidth]{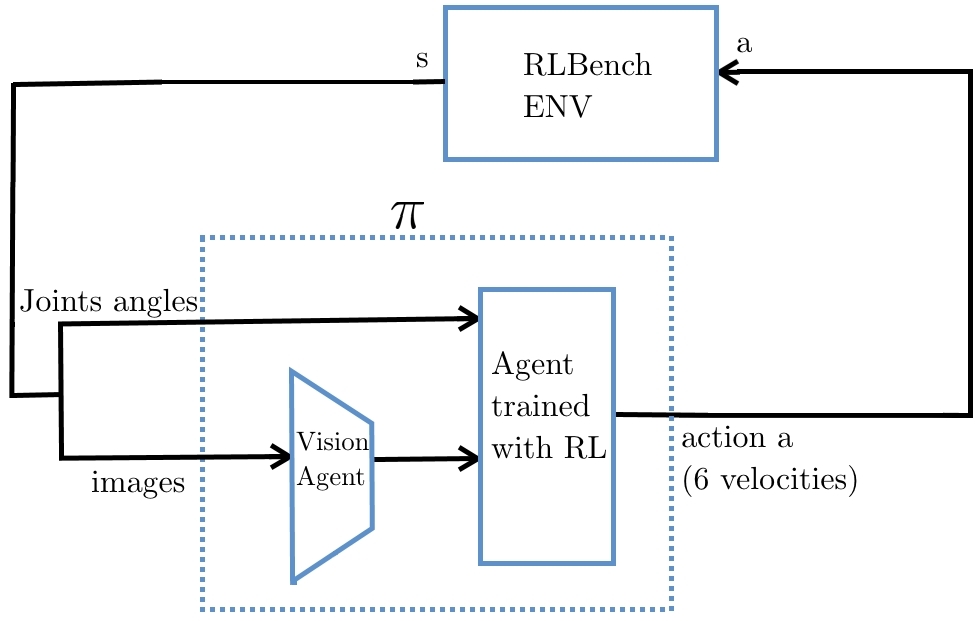}}
\caption{Overview of the training pipeline.}
\label{fig}
\end{figure}

First, the training process starts with the pixel inputs from the camera being analyzed by a computer vision algorithm in order to extract the important features and feed them to the agent. 
The vision based method consisted in 4 different computer vision algorithms to experiment. One is a simple neural network with 2 convolutional layers layers. 2 experiments are focused around the SSD algorithm: one detects only the $(x, y)$ pixel coordinates of the ball and the other computes the distance in the picture between the head and the target. In order to adapt the algorithm to our problem, we created and annotated our own data-set of robot head and target. We then used an SSD model pre-trained on the COCO data-set and fine tuned it on ours. The last is a contrastive learning algorithm that learns high-level features from the images then, like the others vision methods, feeds it to the RL agent.

\section{Background}

Our method is based on the continuous reinforcement learning mechanism, using the SAC as the basic framework. Hence, we first introduce the background of RL and SAC, followed by an explanation on the vision algorithm we used. Then, the overall approach and pipeline is presented in detail.

\subsection{Deep Reinforcement Learning}
Object reaching is a control problem in a stochastic environment where the robot acts by choosing actions sequentially over sequences of time steps $t$, and the goal is to maximize a cumulative reward. We can formalize this RL reaching problem as a Markov decision process (MDP). This process can be modeled as a tuple $(S, A, P, R, \gamma)$, where $S$ and $A$ denote the state and action space respectively, $P$ is the transition probability matrix. $P := \mathbb{P}[s_{t+1} = s′|s_t = s,a_t = a]$ representing the probability distribution to reach a state, given the current state and taking a particular action, $R$ is the state-action associated reward space and can be pictured as $r_t$, and $\gamma \in [0,1]$ is the discounted factor.

The goal of the robot is to learn a policy $\pi_{\theta}$ by interacting with the environment, the policy (i.e., the robot controller) maps $S \rightarrow A$ and is used to select actions in the MDP. It is the “deep” part in DRL as it is commonly parametrized as a Deep Neural Network where $\theta $ is the general parameter storing all the network’s weights and biases. In the case of SAC which is an actor-critic algorithm, 3 neural networks are used. One for the policy and two for the critic. 
The policy can be represented as $\pi(a_t \vert s_t)$, which guides the agent to choose action $a_t$ under state $s_t$ at a given time step $t$ to maximize the future $\gamma$-discounted expected return $R_t = \mathbb{E}[\sum^{\infty}_{i=0} \gamma^i r_{t+i+1}]$.

We performed this learning through the Soft Actor Critic algorithm.

\bigbreak
\textbf{Soft Actor Critic}

We started using the TD3 \cite{fujimoto_addressing_2018} algorithm, but we failed to achieve any sign of learning with it, so we switched to the SAC \cite{haarnoja_soft_2018} algorithm, because it is currently one of the most reliable reinforcement algorithm.    

Instead of solely seeking to maximize the lifetime rewards, SAC \cite{haarnoja_soft_2018} seeks to additionally maximize the entropy of the policy. It brings several advantages: first, it explicitly encourages exploration of the state space and it allows the policy to capture multiple modes of good policies, preventing premature convergence to bad local optima.

It changes the RL problem to : 

\begin{equation}
     \arg \max\limits_{\substack{\pi}} \mathbb{E}_{\pi}[\sum_t \gamma ^t ( r(s_t,a_t) - \alpha \log(\pi(a_t \vert s_t)))]
\end{equation}

where $\alpha$ is the non-negative temperature parameter. 

\subsection{Computer Vision}
In order to rely only on observations from the cameras, a Computer Vision algorithm extracts the features from the RGB pixels. We first tried with a simple 2 layer Convolutional Neural Network (CNN) however this was not sufficient. We then implemented two Deep Learning algorithms: a Single Shot Detector and a contrastive learning algorithm. 
\bigbreak
\textbf{Single Shot Detector}

\noindent The Single Shot Detector (SSD) is a Deep Learning algorithm based on a feed forward convolutional network that produces bounding boxes around areas of interest and then applies a non max suppression to eliminate the ones that have a low confidence score\cite{liu_ssd_2016}. As seen on Fig. 5, the backbone of our used SSD is a MobileNet architecture\cite{howard2017mobilenets}. This has the advantage of being quick and efficient while still achieving high performance. 

The SSD algorithm was chosen over the YOLO algorithm\cite{redmon2016look} for its capacity to be more precise especially for small objects. This was a requirement for our research since we had to detect a small ball in front of the robot. This increased precision of the SSD is explained by its architecture. Indeed, since the SSD applies more convolutional layers to the feature maps of the backbone and uses the outputs of all layers; 
\begin{figure}[htbp]
\centerline{\includegraphics[width=0.5\textwidth]{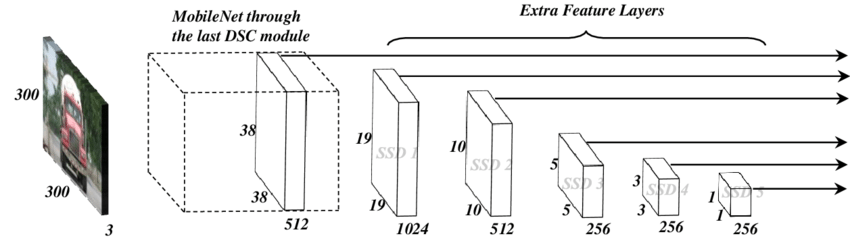}}
\caption{MobileNet-SSD architecture.}
\label{fig}
\end{figure}
the early layers bearing a smaller receptive field are more capable to detect smaller objects of interest\cite{Liu_2018_ECCV}. 
This difference with the YOLO architecture is a needed advantage for us. 

The loss used for training is the original loss from the paper which is a  is a weighted sum of the localization loss (loc) and the confidence loss (conf) and N the number of matched default boxes\cite{liu_ssd_2016}:
\begin{equation}
    L(x, c, l, g) = \frac{1}{N}(L_{conf} (x, c) +\alpha L_{loc}(x, l, g))
\end{equation}
The confidence loss is a soft max loss of classes of confidence (c) and the localization loss is a L1 loss between the ground truth boxes (g) and the predicted ones (l). Finally (x) is the indicator for matching the ground truth box to the default box. 
\bigbreak
\textbf{Unsupervised Representation learning: Contrastive Learning}

\noindent In the case of computer vision, contrastive learning is a self-supervised method in order to maximize the agreement between two similar images and to minimize it between two different images by learning features. It is an approach to find similar and dissimilar data for a Machine Learning model. Contrastive learning in aid of model-free RL has experienced a recent resurgence since researchers have achieved state of the art results by incorporating it to the RL model.  
Contrastive learning relies on queries ($q$) and keys ($k$). Given a query q and a key k $\in {\mathbb{K}}=\{k_1, k_2, ...\}$, if $k$ is a positive $k_+$(similar to $q$), the goal is to maximize a similarity function $f_{sim}$ if $k$ is a negative $k_-$(different than q), $f_{sim}$ needs to be minimized. In our case, $f_{sim}$ is the dot product  $q^Tk$. Other similarity functions exist such as the bi-linear product $q^TWk$ or euclidean distances.
The loss function used to learn embeddings is the InfoNCE loss\cite{oord2019representation}: 
\begin{equation}
    L_q = \frac{exp(q^TWk_+)}{exp(q^TWk_+)+\sum_{p=0}^{K-1} exp(q^TWk_p)}
\end{equation}

\begin{figure}[htbp]
\centerline{\includegraphics[width=0.5\textwidth]{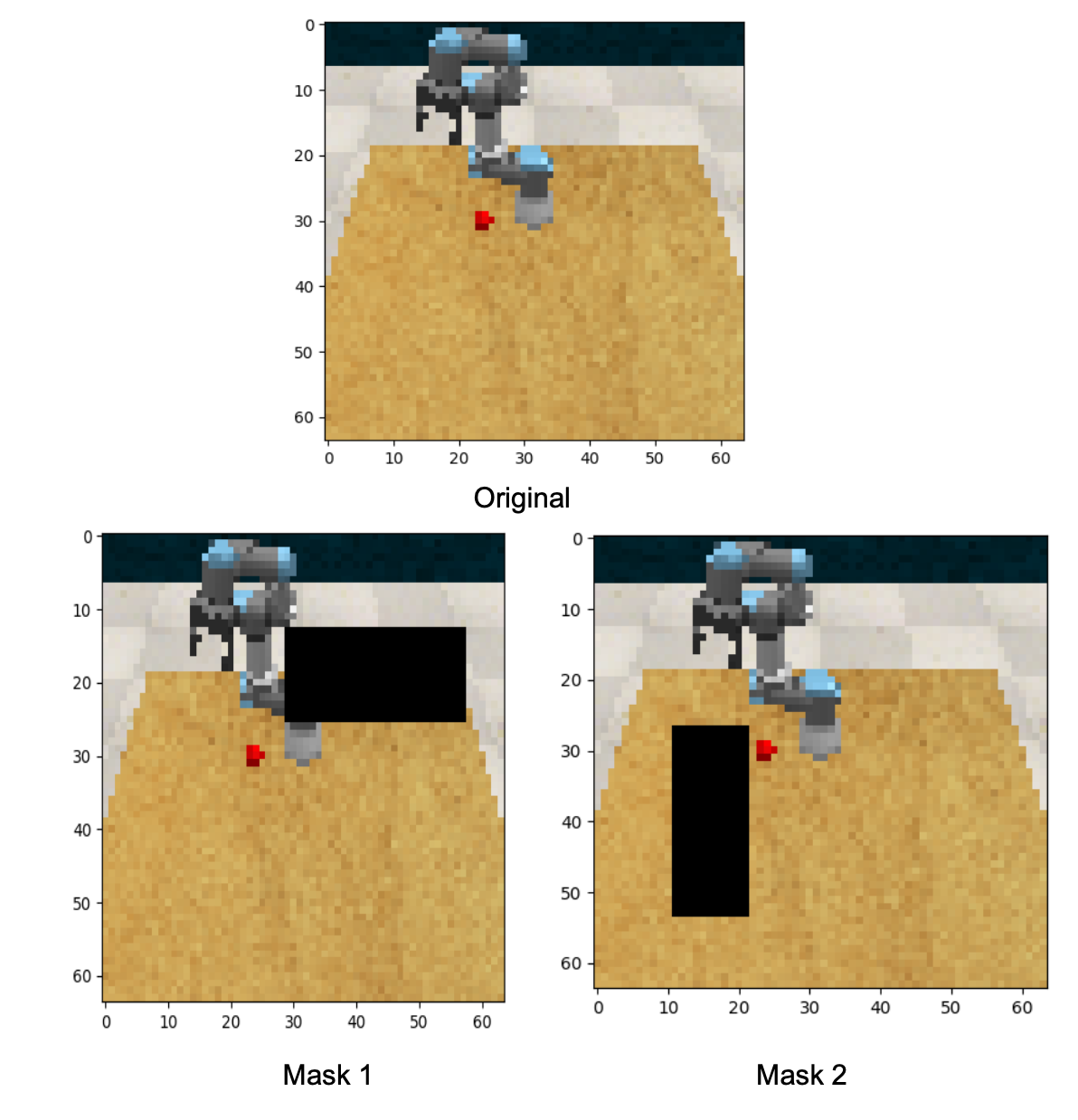}}
\caption{Illustration of data augmentation used for the training of our contrastive learning agent. Mask 1 and 2 corresponds to the anchors used.}
\label{fig}
\end{figure}

State of the art RL from pixels methods use what is called instance discrimination\cite{he2020momentum}. Instance discrimination generates keys and queries as two data augmentations of the same image, negative observations come from an other image. In this paper we use random crop data augmentation, where a random square patch is cropped from the original rendering as shown on Fig. 6, since it has achieved impressive results in computer vision.

\subsection{The overall pipeline}

The overall approach combines the Vision Algorithm and the Reinforcement Algorithm SAC.
 
Here is an overview of the different pipeline used during our vision based experiments. 

\begin{itemize}
    \item \textbf{The first pipeline} is based on the use of our SSD network. Each five steps, the RGB pixels from the front camera are passed through the SSD which predicts two bounding boxes, respectively for the robot's head and the ball-target. If those two boxes are successfully predicted, the normalized distance between the center of the two boxes is computed and added to a buffer containing the 5 most recent distances. If they are not predicted, nothing is added to the buffer in order not to confuse the agent.

    This buffer is concatenated with the current angular positions of the robot's joints to form an observation vector with 11 coordinates, these observations are the states $s_t$ in the MDP, we can therefore write the state space $S$ as $[0,1]^5 \times [-\pi,\pi]^6$.
    \item \textbf{The second pipeline} is very similar to the first since it also uses our SSD network. However, it only predicts the $(x,y)$ pixel coordinates of the ball on the front camera. Therefore, the observation buffer is a concatenation of the pixel coordinates of the ball and the joints' positions of the robot. These coordinates are predicted at the start of each episodes, once they are known, we stop to feed forward the images in the computer vision algorithm. Indeed, since the camera is still and the ball does not move, the coordinates are the same through the episode. 
    
    In this case, $S$ = $[0,1]^2 \times [-\pi,\pi]^6$
    
    \item \textbf{The third pipeline} uses a simple 4 layers neural network to extract features from the front camera and the tip camera. The network consists of 2 convolutional layers and then 2 dense layers. The 2 images are concatenated by channels which results in a (64, 64, 6) image.  
        \begin{figure}[htbp]
\centerline{\includegraphics[width=0.48\textwidth]{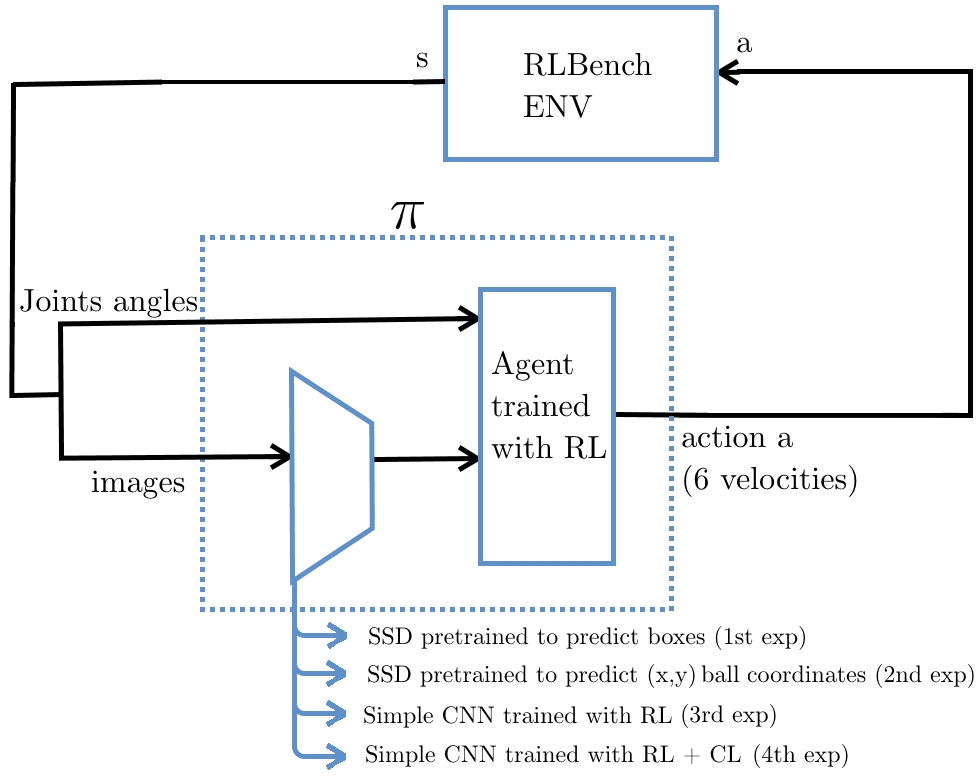}}
\caption{The overall pipeline with the four configurations.}
\label{fig}
\end{figure}
    The output of this network is a vector of $\mathbb{R}^{3000}$ and will be concatenated with the proprioceptions observations of the robot to become the new state in input of the agent.
    \item \textbf{The fourth pipeline} uses a contrastive learning agent to extract high level features from the RGB pixels of the tip and front cameras. It is a 3 layers network: 2 convolutional layers and 1 dense layer with an output of dimension equals to 50. 

    As the precedent experiments, the feature vector is then fed into the agent with its proprioception observations. 
\end{itemize}
The action $a_t$ represents the velocities at each joints as we are controlling the robot through it, we can therefore write $A = \mathbb{R}^6$. All the pipelines are shown on Fig. 7.

\section{Experiments}
In this section, we experiment different combinations of possible pipelines in order to find the most efficient one. We first start by only studying the reward inside the simulation with the ball's and agent's coordinates. This "cheating" method allows us to choose the best reward for our future experiments with a vision based method. Afterwards, we explain the vision based experiments with 3 main ideas based only on the camera stream with a Deep Learning algorithm that analyses this stream. The first uses the predicted bounding box around the head and the ball. The second only predicts the $(x,y)$ pixel coordinates of the ball still with the SSD network at the start of an episode. The third one uses the 4 layers network to extract features from the pixels then feeds to the agent. Finally, we also studied the impact of imitation learning on the agent's training. We then elaborated a final benchmark of all the methods that we experimented in order to have the most sample efficient pipeline. 
Then we try to use contrastive learning to make our agent learn. 

\textbf{Baseline for Benchmark}: 
Every experiments were done with the UR3 6 DOF robot within the RLBench simulator. The metric used is the success rate over the past episodes. It is important to understand that the first experiment to study the reward was done using the ball's and agent's coordinates but all the others were done only with a vision based method. 

\subsection{Experimental Setup}

\begin{figure}[htbp]
\centerline{\includegraphics[width=0.5\textwidth]{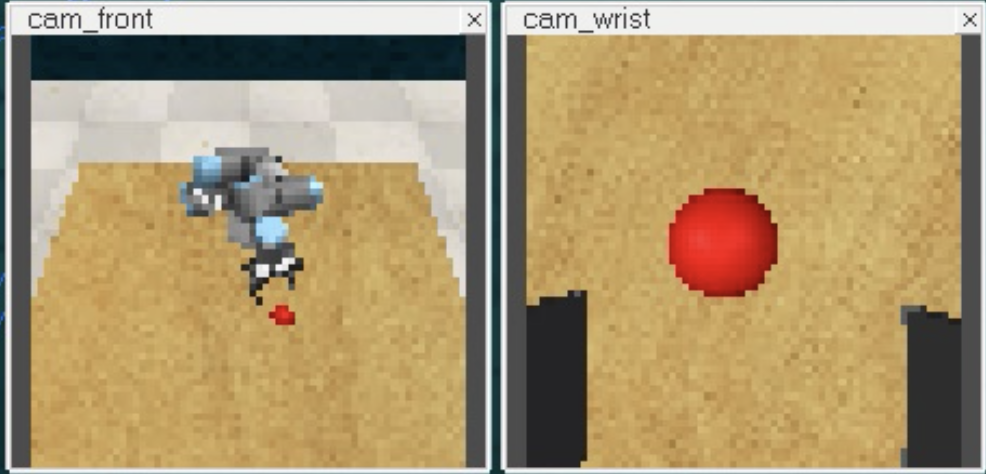}}
\caption{Point Of View (POV) from the two defined cameras. On the left is the front camera and on the right is the wrist camera}
\label{fig}
\end{figure}

To perform our experiments, we set up an RLBench environment for the reaching task. As said earlier, the robot used is the UR3 6 DOF robot.
In the virtual scene, we defined two cameras, one attached to the robot's gripper and the other to the front of the scene (Fig. 8), they will be used during the vision-based method.
Each one of these cameras return a $64 \times 64$ image.

On the algorithm side, we fine tuned the hyperparameters as shown in the following table

\begin{table}[htbp]
\centering
\resizebox{0.35\textwidth}{!}{%
\begin{tabular}{@{}ccc@{}}
\multicolumn{3}{l}{SAC Training Hyperparameters} \\ \midrule
Hyperparameters          & Symbol      & Value   \\
Discount factor & $\gamma$&$0.95$\\
 Temperature & $\alpha$ & $0.2$   \\
Temperature learning rate & $\mu_{\alpha}$ & $10^{-4}$\\
Target update ratio      &  $\tau$     &    $ 0.005$ \\
Actor learning rate      &  $\mu_Q$    &   $ 0.005$\\
Critic learning rate     &  $\mu_{\pi}$&   $ 0.005$\\
Memory buffer size       &  $B$        & $10^{5}$ \\
Batch size               &  $N$        &   $32$   \\
Number of episodes       &  $M$       &$3.10^{4}$\\
Steps per episode        &  $T$        &    $100$   \\
\bottomrule
\end{tabular}%
}
\end{table}

\subsection{Reward Benchmark}
The $1^{st}$ experiment was to study the impact of different rewards on the learning speed. We studied mainly 2 rewards: 
\begin{itemize}
    \item A \textbf{sparse reward} that is close to the "natural" reward we would use for humans. It consists of giving a reward only at the end of an episode, 1 if the agent touched the target and 0 if not. Here, $\epsilon$ is the radius of the ball, $s$ the agent's position and $s_g$ the target's position:
    $$r(s, a) = \left\{
    \begin{array}{ll}
    1, \mbox{if} \|s-s_g\|_2 < \epsilon \\
    0, \mbox{else}
    \end{array}
    \right.
    $$
    \item A \textbf{continuous reward} that returns to the agent a reward at each step instead of each episode. We give him the exponential of minus his distance to the target with a minus in front in order to make him maximize it. If he touches the ball, a bonus of 100 is given:
    
    $$r(s, a) = \left\{
    \begin{array}{ll}
    \exp(-\|s-s_g\|_2) + 100, &\mbox{if } \|s-s_g\|_2 < \epsilon \\
    \exp(-\|s-s_g\|_2)  \mbox{, else}
    \end{array}
    \right.
    $$
    
    We access the distance between the two objects within the virtual environment, nevertheless, it's not a cheating configuration because the reward function is only used during training, and we intend to carry out the training only in simulation.
    
\end{itemize}

After more than 18k episodes, the robot hadn't learned anything with the sparse reward and couldn't reach the ball. However with the continuous reward, he achieved great results of more than 99.5\% of success rate as shown on Fig. 9.

\begin{figure}[htbp]
\centerline{\includegraphics[width=0.5\textwidth]{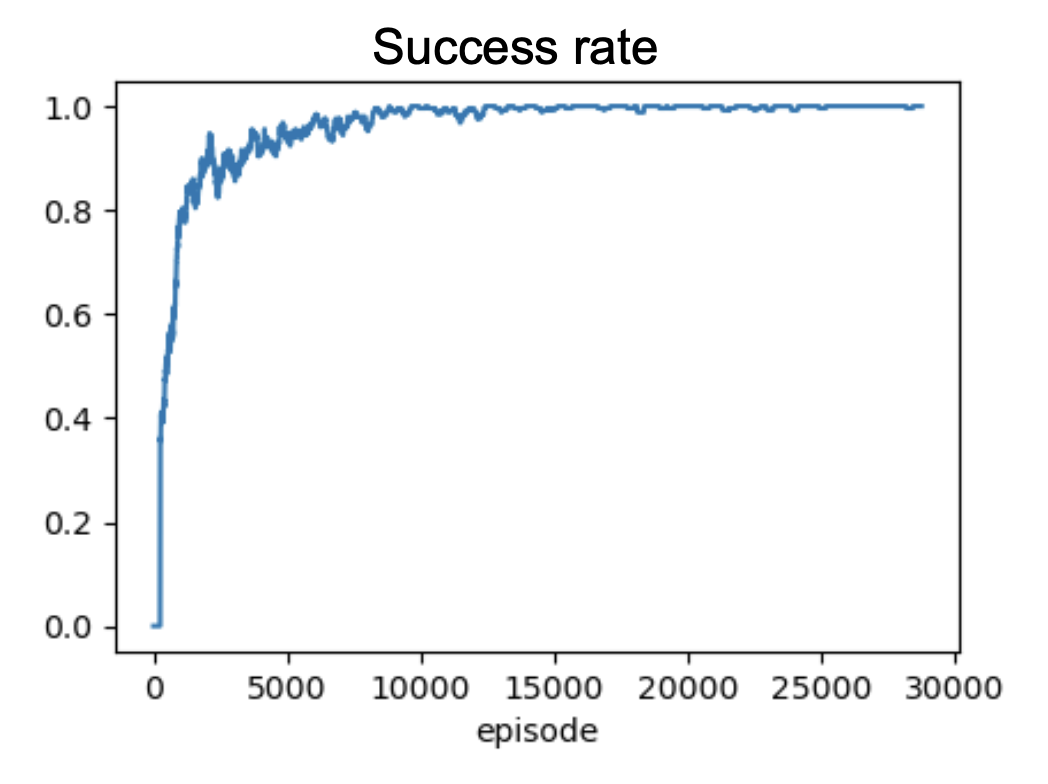}}
\caption{Success rate by episodes for the continuous reward using the ball's and agent's coordinates.}
\label{fig}
\end{figure}

\subsection{Computer Vision experiment}
Now that we know which reward is the most efficient one, we can proceed by studying which vision algorithm works better. The reward used for the next experiments is the continuous reward like for all other experiments. 
An important precision is to understand that giving this continuous reward is necessary and does not affect at all the agent if implemented in real life. Indeed, because of material constraints, we cannot train with the real robot. It is why we decided to train in simulation and to compute a working policy that is then put into the real robot. Therefore, during the training process in simulation we can use the coordinates as a \textbf{reward, not as an observation} to find the most efficient policy without cheating like in the previous reward benchmark. The $2^{nd}$ experiment used the predicted bounding boxes from the SSD algorithm to compute the pixel distance between the target and the agent. This observation was concatenated into a vector with the proprioception of the robot which are his joints' positions. This vector was the input of the DRL agent. The results are given by Fig. 10. We can see that the agent is learning but at a very slow rate. Indeed, this took 7 days of training which is too long to be practical. 

\begin{figure}[htbp]
\centerline{\includegraphics[width=0.5\textwidth]{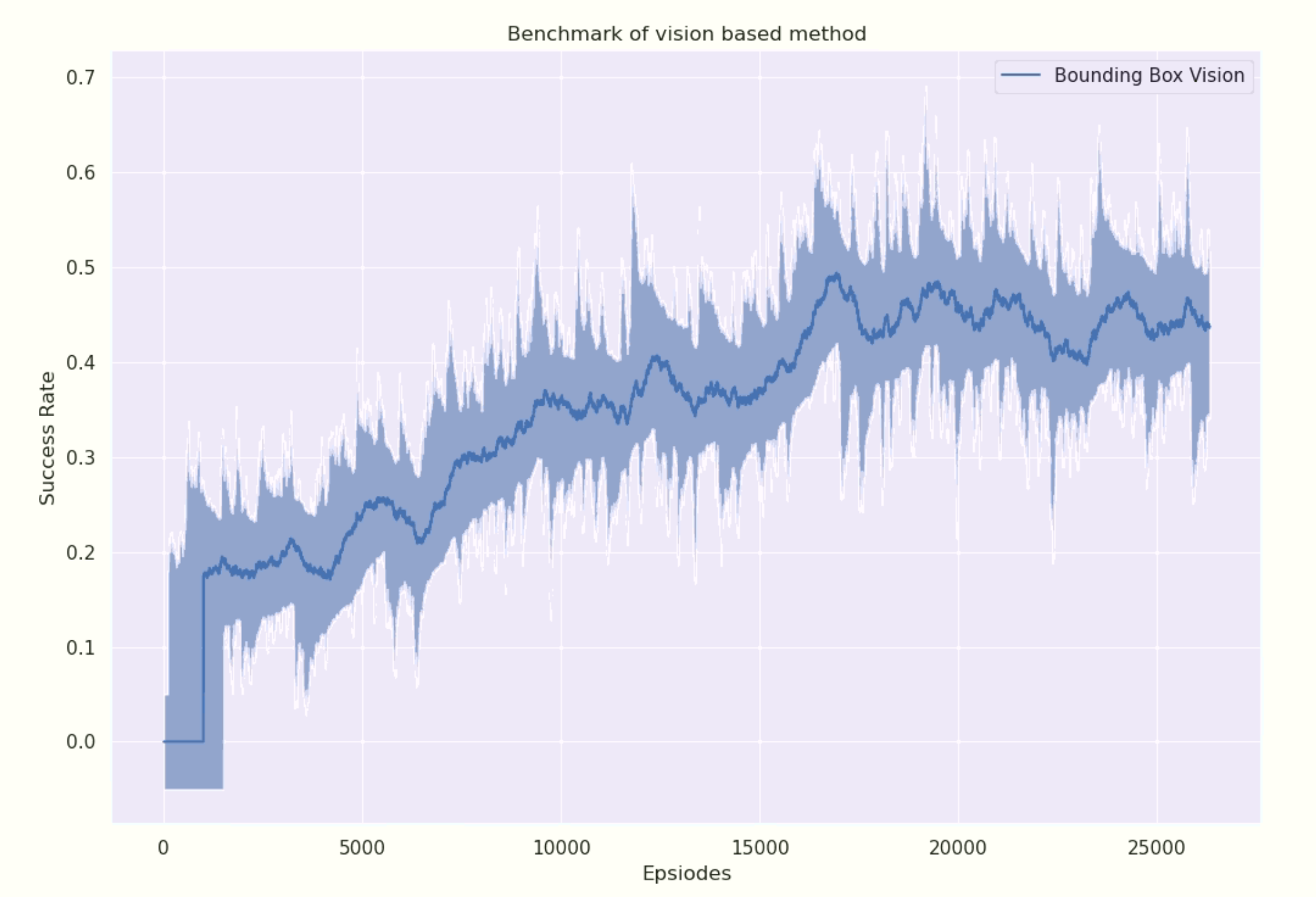}}
\caption{Success rate by episodes for bounding box method.}
\label{fig}
\end{figure}

Since the precedent experience takes too much computational power and is not practical, we decided to try a $3^{rd}$ experiment by locating the $(x,y)$ coordinates of the ball on the image and feed it to the agent with its proprioception information. 
We can see on Fig. 11 that the results are much better: the agent learns better and faster which is convenient. 

\begin{figure}[htbp]
\centerline{\includegraphics[width=0.5\textwidth]{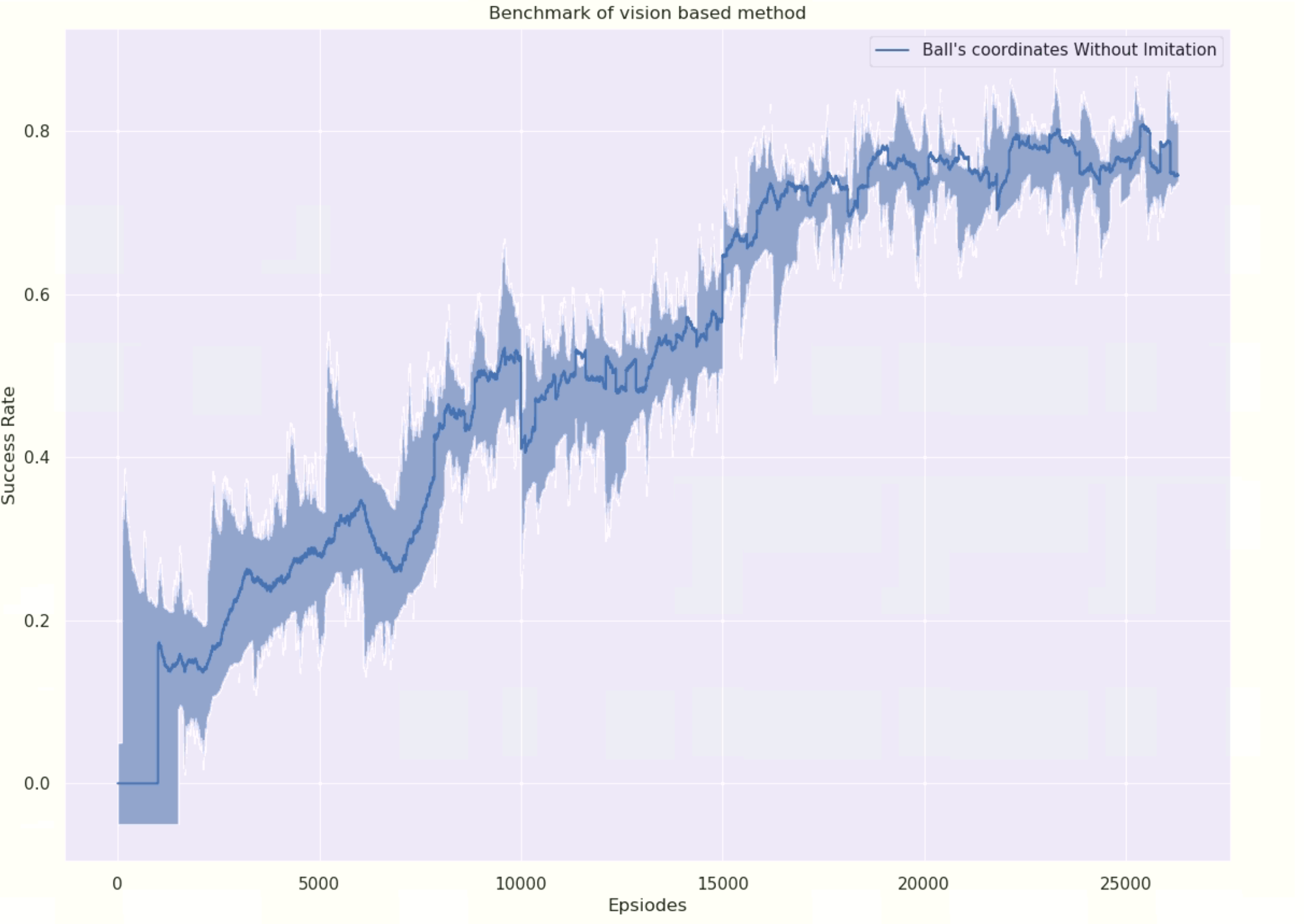}}
\caption{Success rate by episodes for pixel coordinates method.}
\label{fig}
\end{figure}

Then we have the $3^{rd}$ experiment using a simple convolutional neural network upstream of the agent to extract features and give them as inputs. At first, the results were showing no learning pattern and after more than 8k episodes the agent did not seem to learn. It is why we decided to help him by adding \textbf{Imitation Learning} to the process. Imitation Learning intuitively improves the learning since the robot has demos to take example from. To put demos in the agent's buffer is very useful especially for tasks we know perfectly how to execute such as reaching a target. Fig. 12 illustrates the improvements brought by imitation learning, the agent learns very quickly with a success rate of more than 80\% after 30k episodes compared to the experiment without any imitation.

Since Imitation Learning seems to work very well, we also decided to add it to the $2^{nd}$ experiment using the ball's $(x, y)$ pixel coordinates. We can clearly understand with Fig. 13 the impact of imitation learning and more precisely the importance of it at the start of the training since it accelerates the training for the first episodes. 

Finally, here is our final benchmark of the first 2 experiments using vision based methods, illustrated by Fig. 14. 
\begin{figure}[htbp]
\centerline{\includegraphics[width=0.5\textwidth]{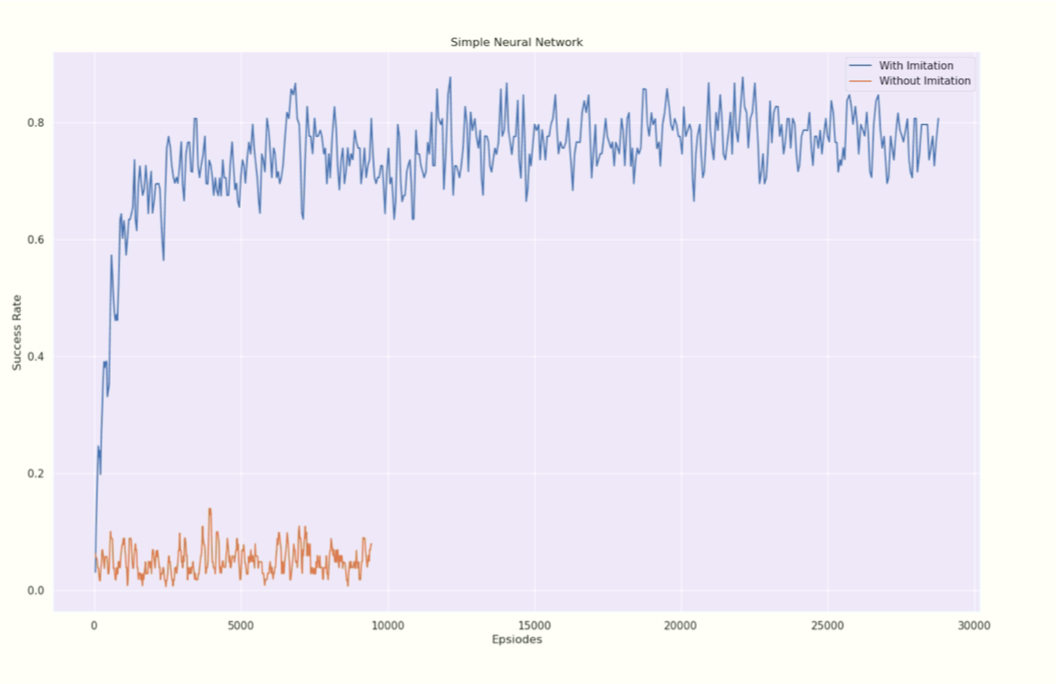}}
\caption{In orange is the success rate by episodes for the training without imitation learning and in blue is the same training with imitation learning.}
\label{fig}
\end{figure}

\begin{figure}[htbp]
\centerline{\includegraphics[width=0.5\textwidth]{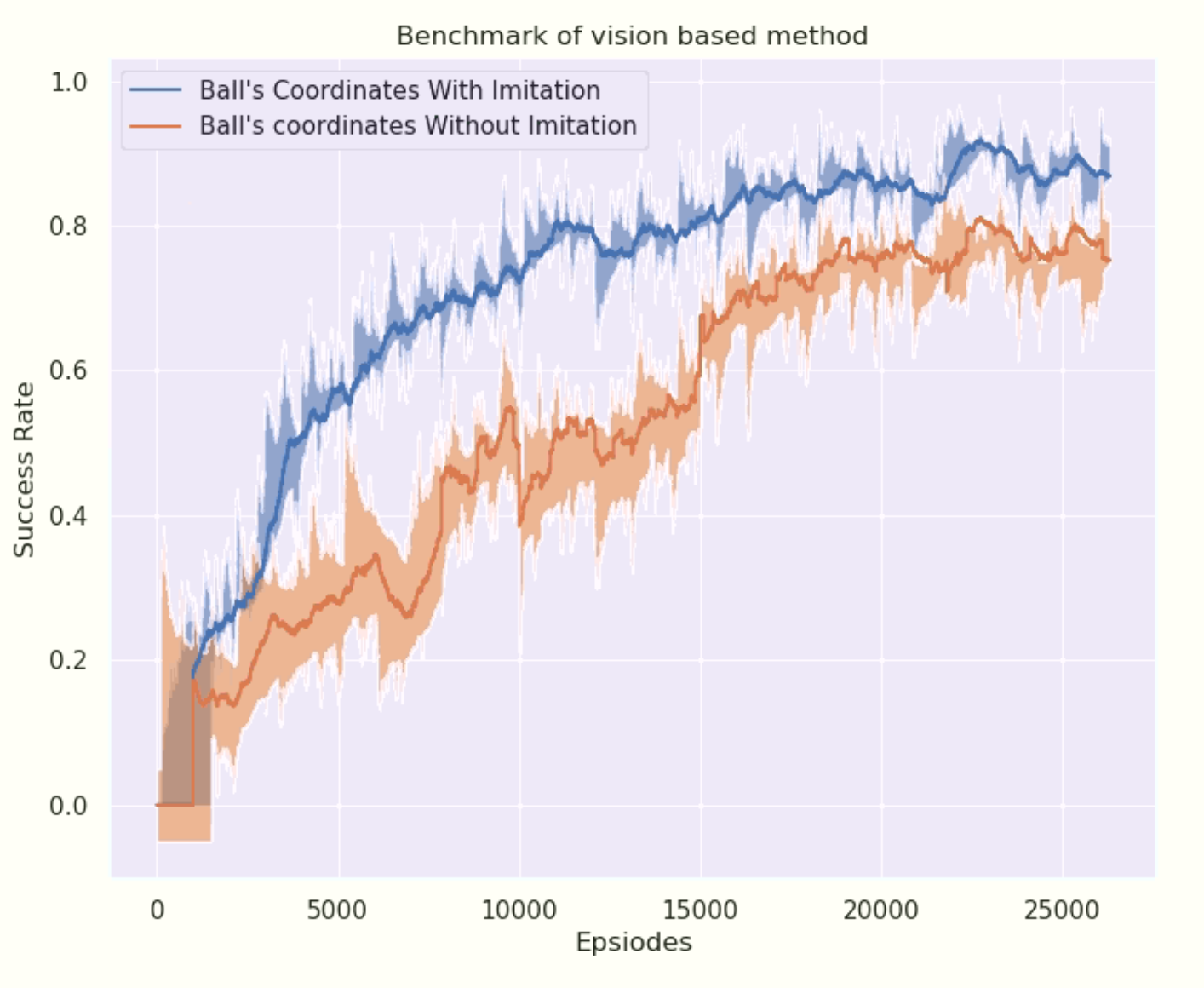}}
\caption{In orange is the success rate by episodes for the training without imitation learning and in blue is the same training with imitation learning.}
\label{fig}
\end{figure}

The 2 most sample efficient are the one using the pixel coordinates of the ball and the one using a simple convolutional neural network. 
Both are using imitation learning to accelerate the training process.

Table 1 offers a wider view of all the performances of all the experiments studied during our research, with the spatial coordinates and with vision only. 

\begin{table*}[]
\resizebox{.9\textwidth}{!}{%
\begin{tabular}{@{}llllllll@{}}
\toprule
                                                 & \multicolumn{7}{l}{\textbf{Overview of successes rates for several configurations}}                                                                                                                                                                                                                       \\ \midrule
\multicolumn{1}{l|}{}                            & \multicolumn{1}{l|}{}                                                                                                 & \multicolumn{3}{l|}{\textbf{No imitation}}                                              & \multicolumn{3}{l|}{\textbf{Imitation}}                                                 \\ \cmidrule(l){2-8} 
\multicolumn{1}{l|}{}                            & \multicolumn{1}{l|}{Configuration \textbackslash Number of episodes}                                                  & \multicolumn{1}{l|}{10k}    & \multicolumn{1}{l|}{20k}    & \multicolumn{1}{l|}{30k}    & \multicolumn{1}{l|}{10k}    & \multicolumn{1}{l|}{20k}    & \multicolumn{1}{l|}{30k}    \\ \midrule
\multicolumn{1}{|l|}{\multirow{2}{*}{No vision}} & \multicolumn{1}{l|}{\textbf{\begin{tabular}[c]{@{}l@{}}Cheating ball coordinates\\ (Sparse reward)\end{tabular}}}     & \multicolumn{1}{l|}{3.5\%}  & \multicolumn{1}{l|}{5.2\%}  & \multicolumn{1}{l|}{6.6\%}  & \multicolumn{1}{l|}{}       & \multicolumn{1}{l|}{}       & \multicolumn{1}{l|}{}       \\ \cmidrule(l){2-8} 
\multicolumn{1}{|l|}{}                           & \multicolumn{1}{l|}{\textbf{\begin{tabular}[c]{@{}l@{}}Cheating ball coordinates\\ (Continuous reward)\end{tabular}}} & \multicolumn{1}{l|}{98.2\%} & \multicolumn{1}{l|}{99.9\%} & \multicolumn{1}{l|}{\textbf{99.9\%}} & \multicolumn{1}{l|}{}       & \multicolumn{1}{l|}{}       & \multicolumn{1}{l|}{}       \\ \midrule
\multicolumn{1}{|l|}{\multirow{3}{*}{Vision}}    & \multicolumn{1}{l|}{\textbf{Distance between two boxes}}                                                              & \multicolumn{1}{l|}{31.3\%} & \multicolumn{1}{l|}{42.7\%} & \multicolumn{1}{l|}{55.4\%} & \multicolumn{1}{l|}{}       & \multicolumn{1}{l|}{}       & \multicolumn{1}{l|}{}       \\ \cmidrule(l){2-8} 
\multicolumn{1}{|l|}{}                           & \multicolumn{1}{l|}{\textbf{Only the ball's box coordinates}}                                                         & \multicolumn{1}{l|}{45.2\%} & \multicolumn{1}{l|}{78.1\%} & \multicolumn{1}{l|}{\textbf{82\%}}   & \multicolumn{1}{l|}{73.1\%} & \multicolumn{1}{l|}{82.2\%} & \multicolumn{1}{l|}{85.2\%} \\ \cmidrule(l){2-8} 
\multicolumn{1}{|l|}{}                           & \multicolumn{1}{l|}{\textbf{Raw image}}    & \multicolumn{1}{l|}{6.1\%}  & \multicolumn{1}{l|}{7.1\%}  & \multicolumn{1}{l|}{6.4\%}  & \multicolumn{1}{l|}{70.6\%} & \multicolumn{1}{l|}{79.9\%} & \multicolumn{1}{l|}{81.3\%} \\ 
\cmidrule(l){2-8} 
\multicolumn{1}{|l|}{}                           & \multicolumn{1}{l|}{\textbf{Raw image + CURL}}    & \multicolumn{1}{l|}{1.4\%}  & \multicolumn{1}{l|}{0.7\%}  & \multicolumn{1}{l|}{2.1\%}  & \multicolumn{1}{l|}{91.2\%} & \multicolumn{1}{l|}{97.6\%} & \multicolumn{1}{l|}{\textbf{98.3\%}} \\ \bottomrule
\end{tabular}%
}
\end{table*} 

\begin{figure}[htbp]
\centerline{\includegraphics[width=0.5\textwidth]{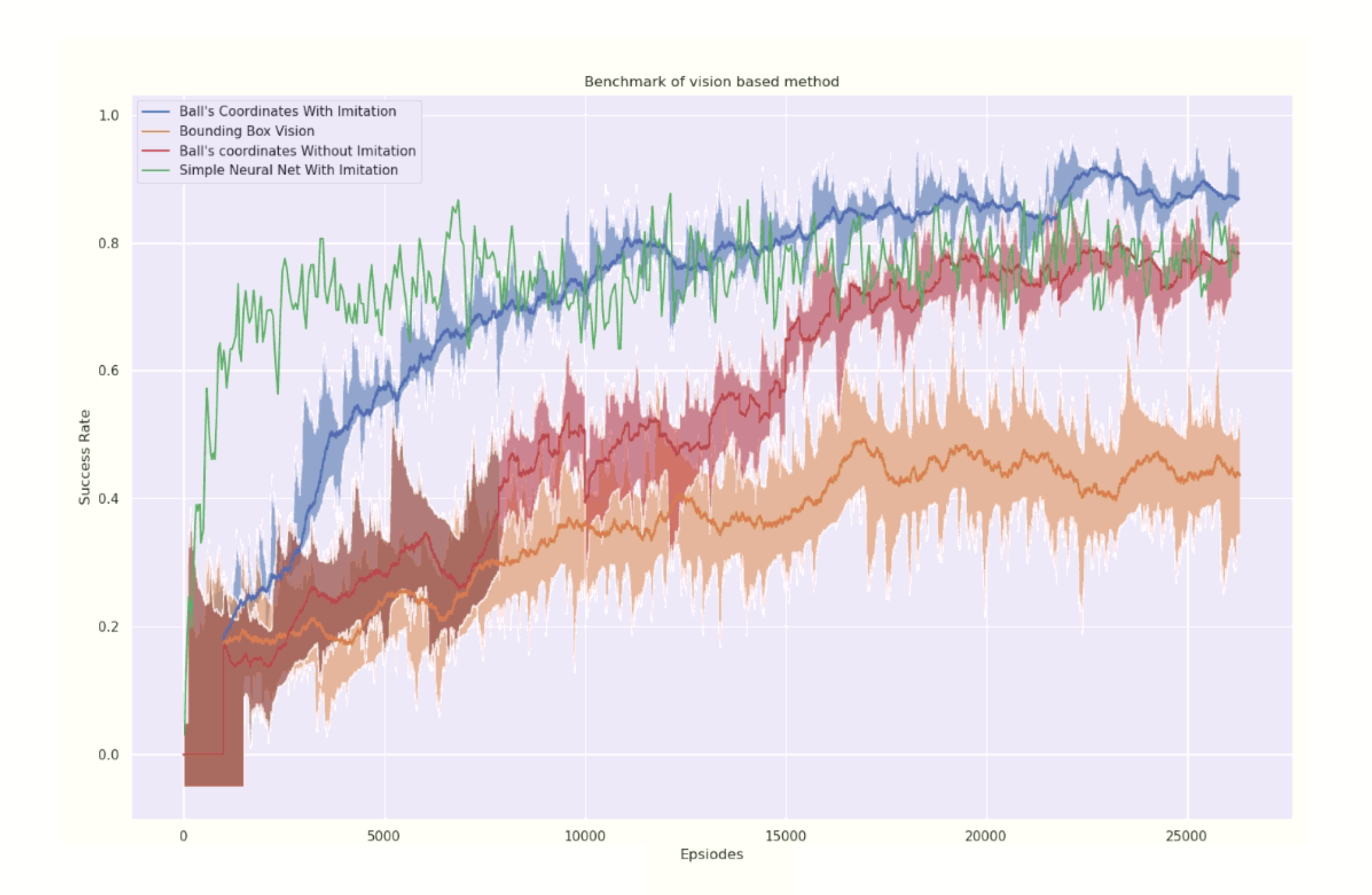}}
\caption{In orange is the success rate by episodes for the training using the bounding box method, in red and blue is the training using the pixel coordinates of the ball respectively without and with imitation and the green plot is the training using the neural network.}
\label{fig}
\end{figure}

Our $4^{th}$ and final experiment was to try the contrastive learning method which reduces the input space and facilitates the upstream vision part since it generalize this method to all use case, not only reach or grabbing objects but also opening doors... We tried with and without imitation learning. The experiment without imitation learning did not work and the results are not worth analyzing. Nevertheless, the one with imitation learning showed impressive improvements compared to the others. In order to illustrates the results of the CURL experiment, we only plotted it with experiment 3. Fig. 15 illustrates that CURL achieves better results with less training episodes. This method is clearly the best method of the whole benchmark. 
\begin{figure}[htbp]
\centerline{\includegraphics[width=0.5\textwidth]{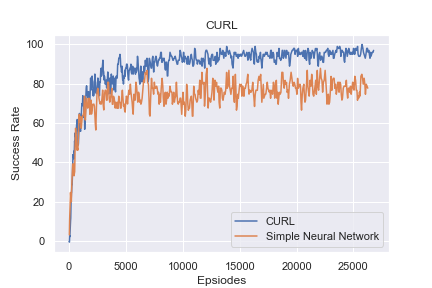}}
\caption{In orange is the results of experiment n°3, in blue is the success rate by episodes for the training using CURL as the encoder.}
\label{fig}
\end{figure}

\section{Conclusion}
In this work we proposed to establish a sample efficient pipeline to train a deep reinforcement agent in vision based robotics. The challenge was to find the most efficient way to train an agent based only on the camera input and the proprioception sensors of the robot. Our final pipeline can be easily deployed to real life robots and can be generalized to many different tasks. We hope that this enables future real life usage of RL robots.

\section{Acknowledgment}
We would like to especially thank Jesùs BUJALANCE MARTIN for his wise advice and his help during our 2 months of research. We also thank Fabien MOUTARDE and MINES PARIS for making this project possible and supervising it.
\bibliographystyle{unsrt}
\bibliography{biblio_dima.bib}

\end{document}